\documentclass{article}
\usepackage{spconf,amsmath,amsfonts,graphicx}
\usepackage{algorithm,algorithmic}
\usepackage{hyperref}
\usepackage{url}

\title{Unsupervised Deep Hashing for Large-scale Visual Search}

\name{Zhaoqiang Xia$^{1}$, Xiaoyi Feng$^{1}$, Jinye Peng$^{1}$, Abdenour Hadid $^{1,2}$}
\address{$^{1}$  School of Electronics and Information, Northwestern Polytechnical University\\ 710129, Xi'an, Shaanxi, China\\
$^{2}$ Center for Machine Vision Research (CMV), University of Oulu, Finland}

\begin{document}

\maketitle

\begin{abstract}
Learning based hashing plays a pivotal role in large-scale visual search. However, most existing hashing algorithms tend to learn shallow models that do not seek representative binary codes. In this paper, we propose a novel hashing approach based on unsupervised deep learning to hierarchically transform features into hash codes. Within the heterogeneous deep hashing framework, the autoencoder layers with specific constraints are considered to model the nonlinear mapping between features and binary codes. Then, a Restricted Boltzmann Machine (RBM) layer with constraints is utilized to reduce the dimension in the hamming space. Extensive experiments on the problem of visual search demonstrate the competitiveness of our proposed approach compared to state-of-the-art. 
\end{abstract}

\vspace{-2mm}
\section{Introduction}
\label{sec:intro}
\vspace{-2mm}
In the era of big data, large-scale visual search is vital for accessing and processing a huge amount of images and this is of great importance in many fields of computer vision. Compared to tree based approaches, hashing based methods utilize several hash functions to project image features into binary codes and are more suitable for large-scale visual search due to compact representations \cite{Paulev2010Locality,Wang2012Semi}. Therefore, the hashing approaches are becoming appealing for dealing with high-dimensional data.

Broadly speaking, existing hashing approaches can be classified into two categories: \textit{data-independent} methods \cite{Gionis1999Similarity,Slaney2008Locality,Kulis2009Kernelized} and \textit{data-dependent} methods \cite{Wang2012Semi, Weiss2008Spectral,Gong2011Iterative,He2013K,Wu2013Semi}. Data-independent methods basically project the data into a hamming space by random hash functions whereas data-dependent methods (also referred to as \textit{learning based hashing}) usually learn hash functions from training datasets by optimizing an objective function. In the case of data-independent methods, the random projections require many binary bits to achieve good performance. With the growing size of data, these methods tend to suffer from memory constraints. The learning based methods, on the other hand, can learn more discriminative hash functions with different objective functions while the hash bits are not linearly growing with the data size. Hence,  learning based hashing is clearly more suitable for large-scale data and is the focus of recent research.

It appears that most hashing approaches use linear models to map the data into binary codes. Hence, most existing methods using learning techniques do not capture well the nonlinear relationships within images. Although several improvements have been proposed, e.g. by adding kernelization \cite{Chang2012Supervised}, it is still challenging to select an appropriate kernel function for specific data. As deep learning techniques are shown to capture well the nonlinear relationships within data, a deep architecture can effectively boost the learning of hash functions. 

In this paper, we propose a novel deep hashing method to learn hierarchy and nonlinear hash functions for obtaining compact binary codes. Our proposed deep architecture includes two heterogeneous layers: autoencoder layers and an RBM (a Restricted Boltzmann Machine) layer. The autoencoder layers are used to generate the initial binary codes whereas the RBM layer is utilized to reduce the dimensionality of the binary codes. For learning the deep autoencoders and RBM, we introduce new objective functions minimizing the reconstruction error and energy function under the constraints of balanced and uncorrelated bits. Extensive experimental analysis on the problem of large-scale visual search demonstrates the validity and competitiveness of our proposed approach compared to state-of-the-art methods.

\vspace{-4mm}
\section{RELATED WORK}
\label{sec:prior}
\textbf{Learning-Based Hashing.} According to whether the semantic information is used or not, learning based hashing can be divided into three categories: \textit{unsupervised hashing}, \textit{semi-supervised hashing} and \textit{supervised hashing}. Unsupervised hashing approaches do not use semantic information (such as tags) whereas supervised hashing approaches learn hash functions with semantic information. Semi-supervised approaches model the data with labeled as well as unlabeled data. For the first category, the Spectral Hashing (SH) \cite{Weiss2008Spectral}, ITerative Quantization (ITQ) \cite{Gong2011Iterative} and K-Means Hashing (KMH)  \cite{He2013K} used different objective functions with constraints of binarization loss or/and the variance of binary bits. For the second category, the Semi-Supervised Hashing (SSH) by Wang \textit{et al.} \cite{Wang2012Semi} constructed an objective function minimizing binarization loss of labeled data and maximizing the variance of unlabeled data. The approach was later extended by employing nonlinear hash functions \cite{Wu2013Semi}. For the third category, Linear Discriminant Analysis (LDA) \cite{Fua2011LDAHash} and multiple linear-SVMs \cite{Rastegari2012Attribute} were used as hash functions and trained with large margin criterion. While most methods seek a single linear mapping, we propose a new solution based on a deep learning framework to explore the hierarchy and nonlinear hash mapping.

\textbf{Deep Learning.} Recently, several deep learning algorithms have been proposed in machine learning and applied to visual object detection and recognition, image classification, face verification and many other research problems \cite{Lecun2015Deep}. Since several foundational deep learning frameworks, such as Convolutional Neural Networks (CNN) \cite{Krizhevsky2012ImageNet}, Stacked AutoEncoders (SAE) \cite{Vincent2008Extracting} and Deep Belief Network (DBN) \cite{Salakhutdinov2006Reducing}, have been presented, numerous deep learning approaches are developed based on  these frameworks. Some deep learning approaches have been applied for learning binary codes. Liong \textit{et al.} \cite{Liong2015deep} presented a framework minimizing a global quantization loss function with two constraints to learn binary codes. In \cite{Xia2014Supervised, Lin2015Deep, Lai2015Simultaneous}, the convolutional neural networks were utilized to extract visual features and a hashing layer was combined to learn binary codes through supervised learning. In this context, we propose a novel deep learning approach with a heterogeneous architecture and specific constraints for image hashing. In the architecture, we use the layer-wise unsupervised learning to learn the model parameters.

\vspace{-2mm}
\section{Deep Hashing}\label{sec:dh}
\vspace{-2mm}

\begin{figure}[t]
  \centering
  \includegraphics[width=0.8\linewidth]{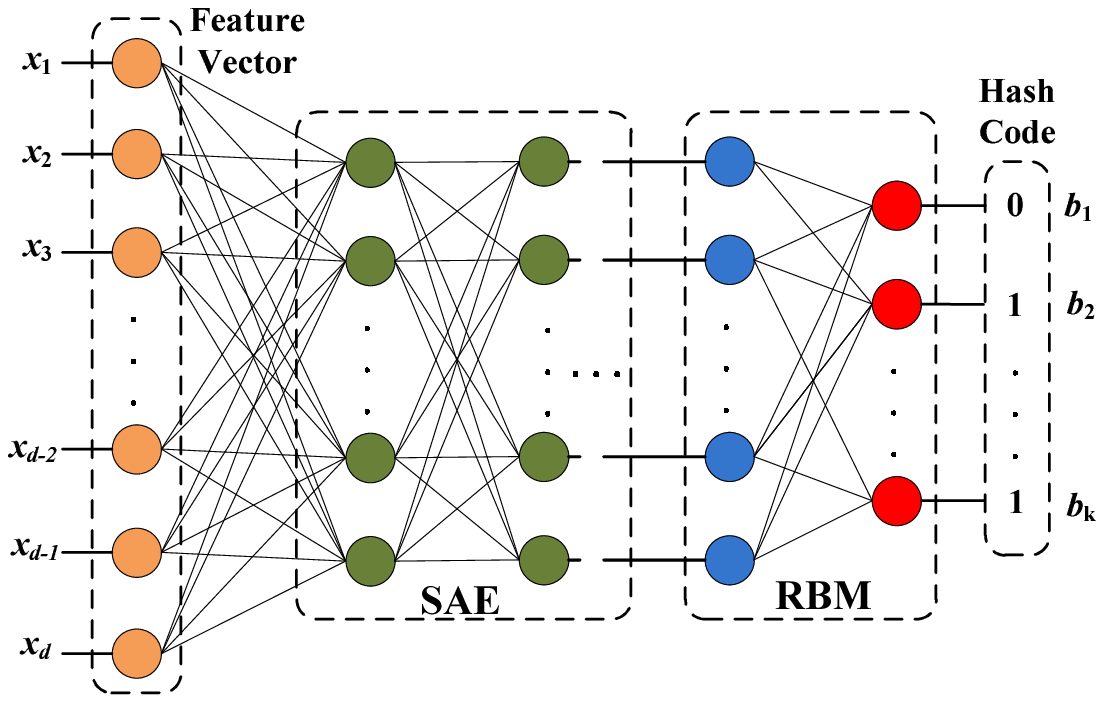}
  \vspace{-2mm}
  \caption{Illustration of our deep hashing architecture.}
  \label{fig:fm}
  \vspace{-2mm}
\end{figure}

Fig. \ref{fig:fm} illustrates the framework of our proposed deep hashing method. The framework contains two heterogeneous layers: (1) several deep autoencoder layers; (2) an RBM layer. Given a feature vector \(\mathbf{x}=(x_1,x_2,...,x_d)^T\), the deep hashing framework can transform the input vector into a binary vector \(\mathbf{b}=(b_1,b_2,...,b_k)^T\), where \(k \ll d \).

\vspace{-4mm}
\subsection{SAE Layers}
\begin{figure}[t]
  \centering
  \includegraphics[width=0.5\linewidth]{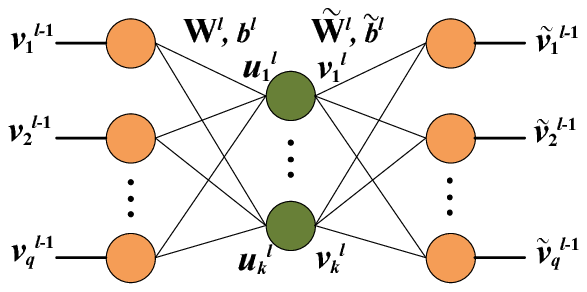}
  \vspace{-2mm}
  \caption{The basic unit of SAE layers for training.}
  \label{fig:sae}
  \vspace{-3mm}
\end{figure}

Let us assume that there are \(L\) layers in our deep autoencoder layers, and the hash function in \(l\)th layer is \(H^l(\mathbf{u}^l)=\tanh(W^l\mathbf{v}^{l-1}+b^l)\). \(\mathbf{u}^l\) represents the input vector in \(l\)th layer and \(\mathbf{u}^0\) is the initial input \(\mathbf{x}\). The output vector in \(l\)th layer is denoted as \(\mathbf{v}^l=(v_1,v_2,...,v_q)^T\). To learn multiple-layers autoencoder, layer-by-layer training has been proposed  \cite{Vincent2008Extracting} to minimize the reconstruction error. As shown in Fig. \ref{fig:sae}, the deep autoencoder (i.e. SAE) can be divided into several three-layers autoencoders for each hidden layer of SAE.  \(\mathbf{\widetilde{v}}\) represents the reconstructed vector of \(\mathbf{v}\). The optimization problem of a conventional autoencoder is to minimize the reconstruction error for each hidden layer: 
\begin{equation}
\min E =\sum_{n=1}^{N}||\mathbf{\widetilde{v}}(n)-\mathbf{v}(n)||_2^2 
\label{eq:sae}
\end{equation}
where \(N\) is the number of training samples.

Besides preserving the similarity in the projected space by minimizing the reconstruction error, the representative hash codes should be balanced and uncorrelated \cite{Weiss2008Spectral}. For a balanced \textit{i}th bit, we should have \(\sum_{n=1}^{N}v_i(n)=0\). In order to be more informative for each bit, the code bits should also be uncorrelated. This is satisfied by setting \(\sum_{n=1}^{N}v_i(n)v_j(n)=0 (i \neq j)\). The solution to the problem (Eq. \ref{eq:sae}) with above constraints is non-trivial as the problem is NP-hard. Since our goal is to obtain the most balanced and uncorrelated bits of hash codes, we propose to add above constraints as regularization terms to seek the suboptimal solution. The regularized optimization problem is defined by
\begin{equation}
\begin{split}
\min_{W^l,b^l} R = &\frac{1}{2}\sum_{n=1}^{N}||\mathbf{\widetilde{v}}^{l-1}(n)-\mathbf{v}^{l-1}(n)||_2^2 + \frac{\lambda}{2}||\sum_{n=1}^N\mathbf{v}^{l}(n)||_2^2 \\
& + \frac{\mu}{2}\sum_{n=1}^{N}||\frac{1}{N}\mathbf{v}^l(n){\mathbf{v}^l}^T(n)-I||_F^2
\end{split}
\label{eq:rsae}
\end{equation}
where \(||\cdot||_F\) represents the Frobenius  norm. \(\mathbf{\widetilde{v}}^{l-1}\) is the reconstructed vector and computed as \(\mathbf{\widetilde{v}}^{l-1} = \tanh(\widetilde{W}^l\mathbf{v}^l+\widetilde{b}^l)\). \(\mathbf{v}^l\) is the output vector of hidden layers and computed as \(\mathbf{v}^l=\tanh(W^l\mathbf{v}^{l-1}+b^l)\).

To learn the model parameters \(\{W^l, b^l\}_1^L\) for  the \(L\)-layers autoencoders, we employ the BackPropagation (BP) algorithm to solve the optimization problem (Eq. \ref{eq:rsae}). \(W^l\) and \(b^l\) are updated as
\begin{equation}
\begin{split}
W^l &:= W^l - \alpha\frac{\partial R}{\partial W^l}\\
b^l &:= b^l - \alpha\frac{\partial R}{\partial b^l}
\end{split}
\label{eq:saeupdate}
\end{equation}
where \(\alpha\) is a learning rate.

The gradients of parameters are derived as
\begin{equation}
\begin{split}
\frac{\partial R}{\partial W^l} &=\sum_{n=1}^{N}{ \mathbf{v}^{l-1}_n\delta^l}^T_n + \lambda\sum_{n=1}^{N}\mathbf{v}^l_n\circ \sum_{n=1}^{N}f'(\mathbf{u}^l_n)\mathbf{v}^{l-1}_n \\
&+\frac{\mu}{N}\sum_{n=1}^{N}(\frac{1}{N}\mathbf{v}^l_n{\mathbf{v}^l}^T_n-I)\mathbf{v}^l_n \circ f'(\mathbf{u}^l_n)\mathbf{v}^{l-1}_n \\
\frac{\partial R}{\partial b^l} &=\sum_{n=1}^{N}\delta^l_n + \lambda\sum_{n=1}^{N}\mathbf{v}^l_n\circ \sum_{n=1}^{N}f'(\mathbf{u}^l_n) \\
&+\frac{\mu}{N}\sum_{n=1}^{N}(\frac{1}{N}\mathbf{v}^l_n{\mathbf{v}^l}^T_n-I)\mathbf{v}^l_n \circ f'(\mathbf{u}^l_n)
\end{split}
\label{eq:pargrad}
\end{equation}
where "\(\circ\)" denotes element-wise multiplication and the sample index "\(n\)" is marked as subscripts for clarify. In Eq. \ref{eq:pargrad}, the local gradient \(\delta^l(n)\) and the derivative of the activation function \(f'(x)\) are computed as
\begin{equation}
\begin{split}
&\delta^l(n) = (\widetilde{W}^l)^T\widetilde{\delta}^l_n \circ f'(\mathbf{u}^l_n) \\
&\widetilde{\delta}^l_n = f'(\mathbf{\widetilde{u}}^l_n) \circ (\mathbf{\widetilde{v}}^{l-1}_n-\mathbf{v}^{l-1}_n) \\
&f'(x) = 1-\tanh^2(x)
\end{split}
\label{eq:delta}
\end{equation}
In Eq. \ref{eq:delta}, the parameters \(\widetilde{W}^l\) and \(\widetilde{b}^l\) need to be learned when the reconstruction errors are back-propagated. These parameters can be learned similarly to the parameters \(W^l\) and \(b^l\).

\subsection{RBM Layer}
\begin{figure}[b]
  \centering
  \includegraphics[width=0.5\linewidth]{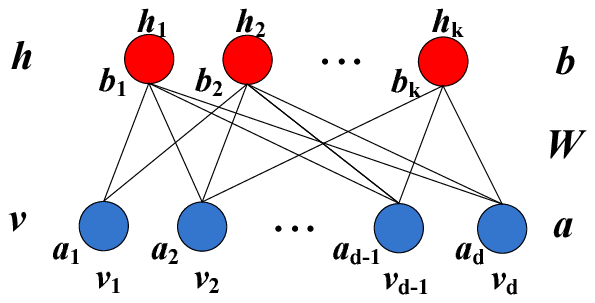}
  \vspace{-2mm}
  \caption{The structure of an RBM layer.}
  \label{fig:rbm}
  \vspace{-3mm}
\end{figure}

We further employ an RBM layer (Fig. \ref{fig:rbm}) to reduce the dimension of the binary codes. Since the variables are binary in the RBM layer, the sign function is used to transform the output vector of SAE layers so that each input unit of the RBM layer can be valued as \(\{0,1\}\).

Let us assume that the visible layer and the hidden layer are denoted as \(\mathbf{v}\) and \(\mathbf{h}\) respectively whereas \(\mathbf{a}, \mathbf{b}\) and \(\mathbf{W}\) are the bias and weights of visible layer and hidden layer. The energy of the RBM model is defined as \(E(\mathbf{v},\mathbf{h})=-\mathbf{a}^T\mathbf{v}-\mathbf{b}^T\mathbf{h}-\mathbf{h}^TW\mathbf{v}\) and the joint probability of \((\mathbf{v},\mathbf{h})\) is \(P(\mathbf{v},\mathbf{h})=\frac{1}{Z}e^{-E(\mathbf{v},\mathbf{h})}\).

The optimization problem of a conventional RBM is to maximize the likelihood of training samples as follows
\begin{equation}
\max  \mathcal{L} = \prod_{n=1}^{N}P(\mathbf{v}(n)) = \prod_{n=1}^{N}\sum_jP(\mathbf{v}(n),\mathbf{h}_j(n)) 
\label{eq:rbm}
\end{equation}
In order to keep the binary codes balanced and uncorrelated, we integrate above constraints into the optimization problem (Eq. \ref{eq:rbm}). Thus, the regularized problem is defined as
\begin{equation}
\min_{W,a,b} J= -ln\mathcal{L}+\frac{\lambda}{2}||\sum_{n=1}^N\mathbf{h}(n)||_2^2  + \frac{\mu}{2}\sum_{n=1}^{N}||\frac{1}{N}\mathbf{h}(n){\mathbf{h}}^T(n)-I||_F^2
\label{eq:nrbm}
\end{equation}
where \(\mathbf{h} = \frac{sgn(W\mathbf{v}+\mathbf{b})+1}{2}\). Since the derivative of sign function is an impulse function, the problem (Eq. \ref{eq:nrbm}) is intractable to compute. To seek the approximate solution, we replace the sign function with a derivable function \(f(x) = \frac{\tanh(\beta x)+1}{2} (\beta \gg 1 )\). Through computing the gradients \(\frac{\partial J}{\partial W},  \frac{\partial J}{\partial b}\) and \(\frac{\partial J}{\partial a}\), the parameters are updated as
\begin{equation}
\begin{split}
W := W - &\alpha\frac{\partial J}{\partial W}\\
b:= b - \alpha\frac{\partial J}{\partial b} &, a:= a - \alpha\frac{\partial J}{\partial a}
\end{split}
\label{eq:rbmupdate}
\end{equation}
We utilize the Contrastive Divergence (CD) algorithm \cite{Fischer2014Training} to seek the  numerical solution of the problem (Eq. \ref{eq:nrbm}).

The gradients are estimated with Gibbs sampling as follows
\begin{equation}
\begin{split}
\frac{\partial J}{\partial W} &\approx \sum_{n=1}^{N}(P_n(\mathbf{h}_i =1|\mathbf{v}^{(r)}_n)\mathbf{v}^{(k)}_n-P_n(\mathbf{h}_i =1|\mathbf{v}^{(0)}_n)\mathbf{v}^{(0)}_n )\\
&+ \lambda\sum_{n=1}^{N}\mathbf{h}_n\circ \sum_{n=1}^{N}f'(W\mathbf{v}^{(0)}_n+\mathbf{b})\mathbf{v}^{(0)}_n \\
&+\frac{\mu}{N}\sum_{n=1}^{N}(\frac{1}{N}\mathbf{v}^{(0)}_n{\mathbf{v}^{(0)}}^T_n-I)\mathbf{v}^{(0)}_n \circ f'(W\mathbf{v}^{(0)}_n+\mathbf{b})\mathbf{v}^{(0)}_n\\
\frac{\partial J}{\partial b} &\approx \sum_{n=1}^{N}(P_n(\mathbf{h}_i =1|\mathbf{v}^{(r)}_n) - P_n(\mathbf{h}_i =1|\mathbf{v}^{(0)}_n) ) \\
&+ \lambda\sum_{n=1}^{N}\mathbf{h}_n\circ \sum_{n=1}^{N}f'(W\mathbf{v}^{(0)}_n+\mathbf{b}) \\
&+\frac{\mu}{N}\sum_{n=1}^{N}(\frac{1}{N}\mathbf{v}^{(0)}_n{\mathbf{v}^{(0)}}^T_n-I)\mathbf{v}^{(0)}_n \circ f'(W\mathbf{v}^{(0)}_n+\mathbf{b}) \\
\frac{\partial J}{\partial a} &\approx \mathbf{v}^{(r)}- \mathbf{v}^{(0)}
\end{split}
\label{eq:rbmgrad}
\end{equation}
where \(\mathbf{v}^{(r)}\) represents the \textit{r}-step Gibbs sampling and \(P(h=1|\mathbf{v}) = sigmoid(W\mathbf{v}+\mathbf{b})\). The derivate of \(f(x)\) is \(f'(x) = \beta-\beta\tanh^2(\beta x)\).

\begin{algorithm}[t]
\caption{The Heterogeneous Deep Hashing (HetDH)}
\label{al:dh}
\begin{algorithmic}[1]
\STATE \textbf{Initialization:} Set up \(\lambda, \mu, \beta, \alpha, L\); Set up iteration times \(T\) and convergence errors \(\epsilon_1, \epsilon_2\); Randomly initialize elements of \(W, b, a \) in \( [0,1]\); Split the training set into \(M\) epochs, each having \(N\) images.
\STATE \textbf{Optimization:} \\
\textbf{for} \(t\) = 1, 2, ..., \(T\)  \\
	\quad \textbf{for} \(m\) = 1, 2, ..., \(M\)  \\
		\quad \; (a) train all layers of SAE successively\\
		\quad \quad \textbf{for} \(l\) = 1, 2, ..., \(L\) \\
		\quad \quad \quad Update parameters \(W^l\) and \(b^l\) by  (Eqs. \ref{eq:saeupdate}, \ref{eq:pargrad}, \ref{eq:delta});\\
		\quad \quad \textbf{end}\\
		\quad \;  (b) train single layers of RBM\\
		\quad \quad Update parameters \(W\), \(b\) and \(a\) by (Eqs. \ref{eq:rbmupdate} and \ref{eq:rbmgrad});\\
	\quad\textbf{end} \\
	\quad \textbf{if} \(1<t<T \)\\
		\quad \quad \textbf{if} \(|{R_t-R_{t-1}|>\epsilon_1}\), \textbf{do} repeat (a) ;\\
		\quad \quad \textbf{if} \(|{J_t-J_{t-1}|>\epsilon_2}\), \textbf{do} repeat (b) ;\\
	\quad \textbf{end}\\
\textbf{end}\\
\STATE \textbf{Return:} All model parameters.

\end{algorithmic}
\end{algorithm}
The detailed deep hashing algorithm is summarized in Algorithm \ref{al:dh}.
\vspace{-4mm}
\section{Experimental Analysis}
\vspace{-2mm}
To evaluate our proposed method, we performed extensive experiments on two datasets: CIFAR-10\footnote{http://www.cs.toronto.edu/~kriz/cifar.html} and MIRFLICKR-25K\footnote{http://press.liacs.nl/mirflickr/}. The CIFAR-10 dataset consists of 60,000 \(32\times32\) color images in 10 classes with 50,000 training images and 10,000 test images. The MIRFLICKR-25K dataset contains 25,000 color images in 26 classes in which 20,000 training images and 5,000 test images are randomly selected. Moreover, the cascaded 512-D GIST \cite{Oliva2001Modeling} and 512-D Bag-of-Features (BoF) \cite{Sivic2009Efficient} are used for image representation.

For comparative analysis, the KLSH \cite{Kulis2009Kernelized}, SH \cite{Weiss2008Spectral}, ITQ \cite{Gong2011Iterative} and KMH \cite{He2013K} algorithms\footnote{The authors have shared their codes on Internet.} are considered and used as baseline methods. Our approach (denoted as HetDH) uses 3 hidden layers (\(600-256-256(128)\) architecture) for SAE and 1 hidden layer (\(16\sim 128\) neurons) for RBM due to the dimensions of the images. To gain insight into the impact of constraints in our proposed deep learning framework, we performed experiments with constraints (denoted as HetDH) and without constraints (denoted as HetDH-WC). We report the results of all the approaches in terms of precision and recall (precision-recall curves). 

\begin{figure}[h]
  \centering
  \begin{minipage}[h]{1.0\linewidth}
  \centering
  \centerline{\includegraphics[width=0.90\linewidth]{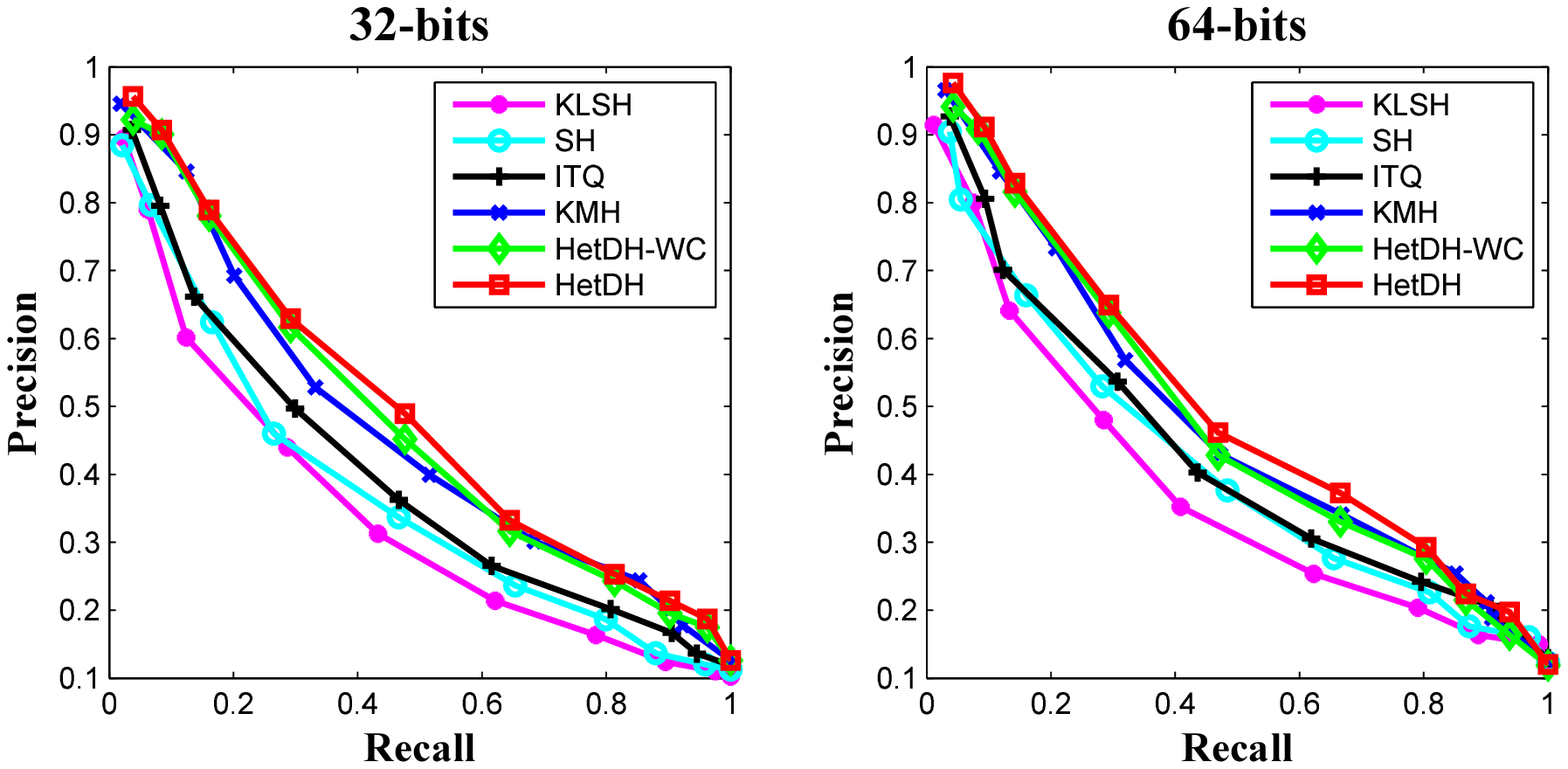}}
  \centerline{(a) CIFAR-10}\medskip
  \end{minipage}

  \begin{minipage}[h]{1.0\linewidth}
  \centering
  \centerline{\includegraphics[width=0.95\linewidth]{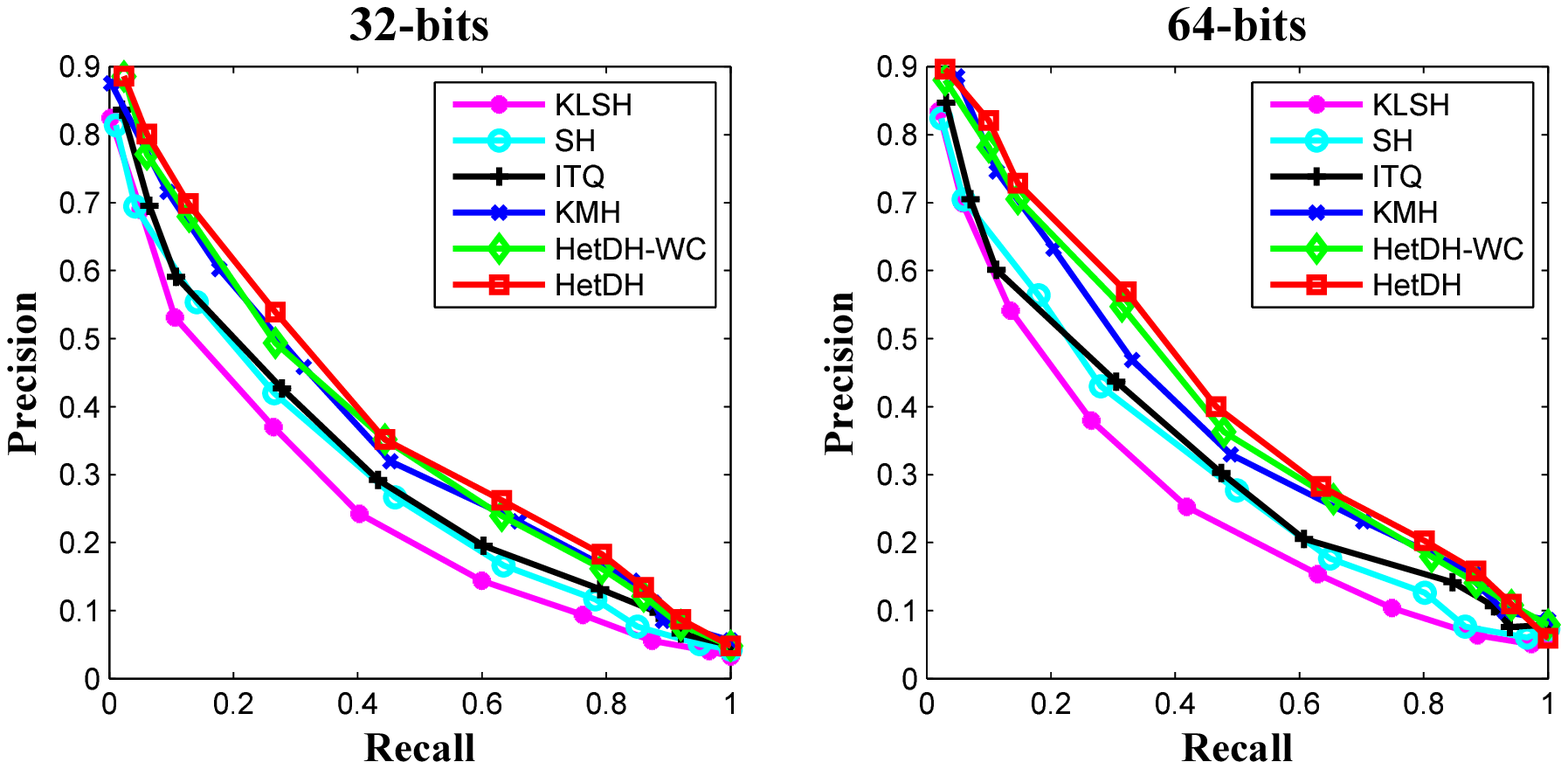}}
  \centerline{(b) MIRFLICKR-25K}\medskip
  \end{minipage}
  \vspace{-2mm}
  \caption{The precision-recall curves on the CIFAR-10 and MIRFLICKR-25K datasets at 32 and 64 bits.}
  \label{fig:results}
  \vspace{-3mm}
\end{figure}
Fig. \ref{fig:results} shows the precision-recall curves on the CIFAR-10 and MIRFLICKR-25K datasets at 32 and 64 bits for all considered methods. It can be observed that our proposed method (HetDH) outperforms all other methods in all configurations. The results also point out that all the learning based methods work better than the data-independent method (KLSH). Compared to shallow learning models (SH, ITQ, KMH), both HetDH and HetDH-WC achieve good performance due to the hierarchy representation of deep learning in our proposed approach. Comparing HetDH with HetDH-WC, the obtained results show that the constraints for each layer effectively boost the conventional deep learning and improve the searching performance. This assesses the effectiveness of our proposed algorithm.

It is worth noting that the dimension of the hash codes (the number of binary bits) affects the performance of image search (see the experimental results at 32 bits vs. 64 bits). The results indicate that larger dimensions improve the precision and recall but at the cost of more memory storage. Finally, the experiments also show that all the hashing methods seem to work better on the CIFAR-10 dataset compared to MIRFLICKR-25K dataset. This is perhaps due to the diverse nature of the images in MIRFLICKR-25K dataset compared to the images in the CIFAR-10 dataset.
\vspace{-3mm}
\section{Conclusion}
\vspace{-2mm}
We proposed a heterogeneous deep learning architecture for learning hash functions. With two constraints for balanced and uncorrelated binary codes, we learned the parameters of SAE and RBM layers. Experimental results and extensive comparative analysis on the problem of large-scale image search assessed the effectiveness of our proposed approach which outperformed state-of-the-art unsupervised methods.

\vfill
\pagebreak

\bibliographystyle{IEEEbib}
\bibliography{refs}

\end{document}